# COMPUTER VISION APPROACH FOR LOW COST, HIGH PRECISION MEASUREMENT OF GRAPEVINE TRUNK DIAMETER IN OUTDOOR CONDITIONS

D. S. Pérez, F. Bromberg, F. Gonzalez Antivilo

The authors are **Diego Sebastián Pérez**, Ph.D. student in the DHARMa lab, **Facundo Bromberg**, Diego's Ph.D. advisor and DHARMa lab director, Department of Information Systems Engineer, Universidad Tecnológica Nacional, Mendoza, Argentina; and **Francisco Gonzalez Antivilo**, Laboratory of Plant Physiology, Department of Agricultural Sciences, Universidad Nacional de Cuyo, Luján de Cuyo, Mendoza, Argentina.
**Corresponding author:** Diego S. Pérez, Rodríguez 273, CP 5500, Mendoza, Argentina; phone: +54 9 (261) 300-8563; e-mail: sebastian.perez@frm.utn.edu.ar

**ABSTRACT.**

*Trunk diameter is a variable of agricultural interest, used mainly in the prediction of fruit trees production. It is correlated with leaf area and biomass of trees, and consequently gives a good estimate of the potential production of the plants. This work presents a low cost, high precision method for the measurement of trunk diameter of grapevines based on Computer Vision techniques. Several methods based on Computer Vision and other techniques are introduced in the literature. These methods present different advantages for crop management: they are amenable to be operated by unknowledgeable personnel, with lower operational costs; they result in lower stress levels to knowledgeable personnel, avoiding the deterioration of the measurement quality over time; and they make the measurement process amenable to be embedded in larger autonomous systems, allowing more measurements to be taken with equivalent costs. To date, all existing autonomous methods are either of low precision, or have a prohibitive cost for massive agricultural adoption, leaving the manual Vernier caliper or tape measure as the only choice in most situations. In this work we present a semi-autonomous measurement method that is susceptible to be fully automated, cost effective for mass adoption, and its precision is competitive (with slight improvements) over the caliper manual method.*

## INTRODUCTION

The main contribution of this work consists in a low cost, high precision measuring method of grapevine trunk diameter based on techniques of Image Processing and Image Segmentation. Trunk diameter is a variable of agricultural interest in the production of fruit trees in general, as trunk carries out with the vital function of transporting water and nutrients from roots to consumption areas (shoots, leaves, fruits), as well as with the accumulation of reserve substances (MacAdam, 2013). Also, it is correlated with the capacity of the plant to sustain growth and maturation of shoots, leaves and fruits, and it is a good estimator of other indicators of the plant productive potential, such as biomass and leaf area (Causton, 1985; Niklas, 1994). There are numerous allometric relationships reported to estimate leaf area and biomass of different trees using mainly the cross-sectional diameter or area of the trunk (Castelan-Estrada, Vivin, & Gaudillière, 2002; Niklas, 1995). In practice, the trunk diameter is measured using manual methods such as Vernier calipers or measuring tapes. Although these instruments can be considered high precision (0.05 mm in case of standard caliper and 0.5 mm in case of standard tape), they may result in important measurement errors. On one hand there are human errors that may occur during extensive measurement campaigns due to fatigue, haste or carelessness (errors I through IV in table 1). Other errors may arise indirectly due to the elevated costs in labor stipends of the manual method that force agronomists to save in the number of trees for which measurements are performed. While, this may be mitigated using interpolation methods such as Krigging (Cressie, 1990), in practice even the minimal number of measurements required by these methods for accurate estimates may be prohibitive. We refer to these as *interpolation errors* and list it together with the other errors in table 1.

Table 1. Potential errors in the manual measuring process based on caliper or tape.

| Type | Description |
|------|-------------|
| I | Reading of the measuring instrument |
| II | Transcribing values into the datasheet |
| III | Writing values in the slot corresponding to its plant ID |
| IV | Transcribing the datasheet into the computer |
| V | Interpolation errors |

Our contribution takes advantage of Image Processing and Image Segmentation techniques. Image Processing is the use of computer methods for preprocessing a digital image and converting it into a form suitable for further analysis (Szeliski, 2010). The term Image Segmentation refers to the process of partitioning a digital image into a set of regions, each of which is strongly correlated with some object class of the real world, or background. The pixels of each class are grouped based on some uniformity criteria (Jain, Kasturi, & Schunck, 1995).

**RELATED WORK**

Several works of autonomous measurement of trunk diameter have been presented previously in the literature. For instance, (Tetuko, Tateishi, & Wikantika, 2001) propose a method for estimating ranges of values of trunk diameters over large areas of dense forests of four species of Java-Indonesian trees. Their approach projects L-band microwaves on the trunks, and uses a Synthetic Aperture Radar (SAR) to read the resulting backscattering coefficient, later correlated with trunk diameter. In their paper they use a SAR mounted on the JERS-1 satellite, which has a resolution of 18 m per pixel, although other existing SARs may range from 200 m per pixel (low resolution) to 1 m per pixel (high resolution). Technologies known as airborne SAR sensors (mounted on airplanes or other atmospheric aircrafts) achieve resolutions of 0.1 m to 0.3 m per pixel, but require specialized and expensive equipment. Satellite SARs, in contrast, require access to a satellite, with lower cost and higher availability than the airborne version, but still prohibitive or inaccessible to most producers. In either case, however, the resolutions are still large enough to prevent their use in any practical PA application. (Jutila, Kannas, & Visala, 2007) present a method designed for forest harvesters. Their method consists in a 2D laser range finder mounted on an all-terrain vehicle, used to obtain depth information. It achieves good precision in pines forests with an error mean of 6 mm and standard deviation of 22 mm. One limitation for adoption of this technique in fruit trees production is the high cost of equipment needed for implementation. In (Omasa, Hosoi, Uenishi, Shimizu, & Akiyama, 2008), a 3D model of a city park was used in order to quantify the biophysical variables of trees, such as height, canopy diameter and trunk diameter. The 3D model was generated with a scanning Light Detection and Ranging (LIDAR) technology, combining data from airborne sensors with data on-ground using portable sensors. The authors reported errors less than 3 mm in the three trees that were measured. While this method shows high precision, the high cost of aerial and portable LIDAR sensors makes it privative for most fruit trees producers. (Kan, Li, & Sun, 2008) presented a method based on Computer Vision for measuring trunk and branch diameters from images acquired by a conventional digital camera. A calibration stick is included in the scene next to the tree trunk, and it is detected using template matching (Brunelli, 2009). The trunks and branches of the tree are also detected by template matching and then the number of trunk and branch pixels is counted. The actual diameters are obtained by multiplying the trunk and branch diameters in pixels by the size of a pixel obtained through the calibration stick. This method was tested on 50 images resulting in a mean error of 6.7 mm and standard deviation of 17.3 mm. However, when the background of the images is complex (for instance, not a clear sky), the calibration stick may not be detected and the method fails to obtain the diameters. The results over the 50 images show a success rate of 90% in detecting the calibration stick. For a definite assessment of their approach one would require information on the proportion of these images with complex backgrounds. This information, however, is not reported. (Thamrin et al., 2013) discussed a new and relatively simple tree diameter measurement technique using a high-performance, non-intrusive infrared sensor. The experiments were conducted in a controlled laboratory environment under ambient light condition produced by a fluorescent lamp, with four cylindrical poles of different diameters used as controlled replacements for trees. About 80% of the experiments show errors of less than 2 mm for each pole, with an average of 2 mm for standard deviations. The rest of the results give errors between 40 mm to 50 mm. Although these results are promising, they still present 20% of cases with unacceptable errors on a controlled environment. All the techniques described above have significantly contributed in the process of obtaining either high precision or low cost technique for measuring the trunk diameter of plants. However, these solutions are not optimal: some are low cost but low precision (or equivalently high failing rate), and others are high precision but high cost. In this work we propose a low cost, high precision method based on Computer Vision for measurement of trunk diameter of grapevines, whose cost is in the order of Kan's technique (the lowest of all) and whose precision is in the order of Omasa's (the highest of all).

Our approach consists in digitally processing images of grapevines obtained with a standard commercial digital camera on field conditions of climate and luminosity. Trunk diameter measurements using image processing present several difficulties. First, the trunk must be segmented out from the other elements of the scene such as soil, leaves, fruits, sky, and other plants. Second, the direction of the measurement must be detected. Third, the preferred measurement position along the trunk must be detected. Fourth, it is necessary to measure the distance in pixels between the edges of the trunk in the direction of measurement just determined, and conclude by re-scaling the measurement from pixels to millimeters. In this work we incorporate the use of a modified quick clamp (shown in figure 1a) to address all these difficulties. The segmentation problem is simplified by changing the color of the spade pads to red as illustrated in figure 1b, a color rarely encountered in vineyards. With this modification it is possible to use a color segmentation technique for a successful and precise identification of the spade pads. The segmentation technique used in this work is explained in the following subsection. With the spade pads segmented as illustrated in figure 1c, all remaining difficulties are easily addressed: the direction of measurement is the same as the perpendicular direction to the segment $CD$ in figure 1c; the position of the clamp along the trunk determines the preferred position for measurement; the trunk diameter is the distance (in pixels) between the internal edges of the spade pads segmented (points $A$ and $B$ in figure 1c); and the re-scaling from pixels to millimeters is performed using the previously known height of the spade pads in millimeters and its measure in pixels (distance between $C$ and $D$ in figure 1c).

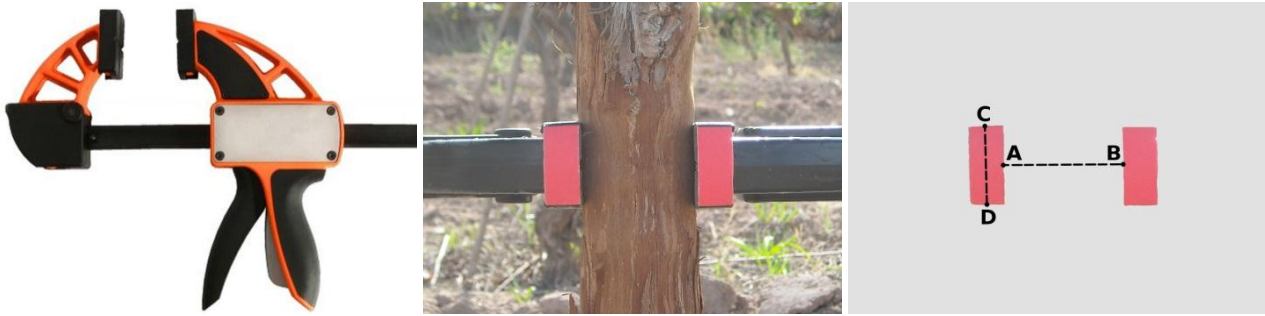

Figure 1. From left to right: (a) Quick clamp; (b) Original image with the quick clamp; (c) Results obtained by the GMM segmentation algorithm for the original image and calculating the distance in pixels between the spade pads (points A-B) and height in pixels of a spade pad (C-D points).

The use of the quick clamp in this work may be justified by its operational simplicity but most importantly because it provides reproducibility to our experiments: being a commercial product it can be universally acquired. But despite this operational simplicity, the clamp used here still requires too much human intervention to competitive with the Vernier caliper: for each trunk to be measured by our method, the measurement process requires of personnel to fix the clamp in the trunk in the preferred position, take the picture, and later run it through our image segmentation method. In practice, however, with no need of reproducibility, the quick clamp can be easily replaced by other more practical clamps (e.g., cheaper, lighter) that could be attached to the plant at the beginning of the season by expert personnel (e.g., researchers), with this cost amortized along several measurements during the whole season. Also, the image processing stage could be easily automated by embedding the image processing algorithm within portable devices such as a cell-phone or tablet that would pipe the photograph taken by its camera to the image-processing software. The design of this alternative clamp and the software for portable devices are technical issues that are beyond the scope of this work, being the image processing method our main focus.

## MATERIALS AND METHODS

This section describes the implementation details of our image-based measurement method and explains the image segmentation algorithm, as well as details on the experimental setups used for evaluation.

### IMPLEMENTATION OF IMAGE-BASED MEASUREMENT METHOD

The first step in the image-based measurement process consists of measuring the distance in millimeters of a pixel, using the previously known height of the spade pads in millimeters and its measure in pixels. Then, the horizontal distance in pixels between the internal edges of the clamps is measured, and finally this distance is multiplied by the distance in millimeters of a pixel to obtain the measure in millimeters. In practice, however, possible segmentation errors on edges of pads were averaged out by measuring several points $A_n$ and $B_n$ to obtain the reported diameter value according to $|A - B|_{pixels} = \frac{1}{N}\sum_{n=1}^{N}|A_n - B_n|$. Finally, the known height of the spade pads in millimeters $|C - D|_{mm}$ and in pixels $|C - D|_{pixels}$, obtained for the segment $CD$ in figure 1c, serves as a re-scaling ratio for converting the diameter $|A - B|_{pixels}$ in pixels to the diameter $|A - B|_{mm}$ in millimeters.

**IMAGE SEGMENTATION ALGORITHM**

The segmentation technique consists of two stages: *training* and *classification*. The training stage uses manually segmented images in separate classes to estimate a probabilistic model based on the color information of the pixels for each class. Later, during the classification stage, the models are used to segment images autonomously by deciding the most probable class of each pixel, given its color.

Let us first introduce the details of the probabilistic model we used: *Gaussian Mixture Models (GMM)* (Bishop, 2009), to later explain in detail the training and classification stages. GMMs are a weighted sum of Gaussian distributions that provide a multi-modal class of multi-variate density models. These models have two important advantages that make them particularly convenient for color image segmentation, as exemplified by the works (Bishop, 2009; Hastie, Tibshirani, & Friedman, 2009; Szeliski, 2010). On one hand, they have strong representational power: by using a sufficient number of Gaussians and by adjusting their means, covariances, and weights of sum, can approximate any density to arbitrary accuracy. On the other hand, there are well-known, computationally efficient techniques for parameters estimation of GMMs. In this work we used the elegant and powerful method of maximum likelihood executed with the Expectation-Maximization (EM) algorithm. EM first chooses random initial values for the model parameters, i.e., means, covariances, and weights of the sum. Then, it alternates between the following two stages: (i) the expectation step, or E step, uses the current values for the parameters to evaluate the posterior probabilities; (ii) then uses these probabilities in the maximization step, or M step, to re-estimate the means, covariances, and weights of sum using techniques such as *maximum-likelihood estimation*. The algorithm proceeds iteratively until it converges, i.e. when the change in the log likelihood function, or alternatively in the parameters, falls below some threshold. The works in (Bishop, 2009; Hastie et al., 2009) explain in detail how to combine maximum likelihood in the algorithm Expectation-Maximization for estimating the numerical parameters of the GMM (i.e., means, covariance matrices, and weights). Let us describe now the *training stage*.

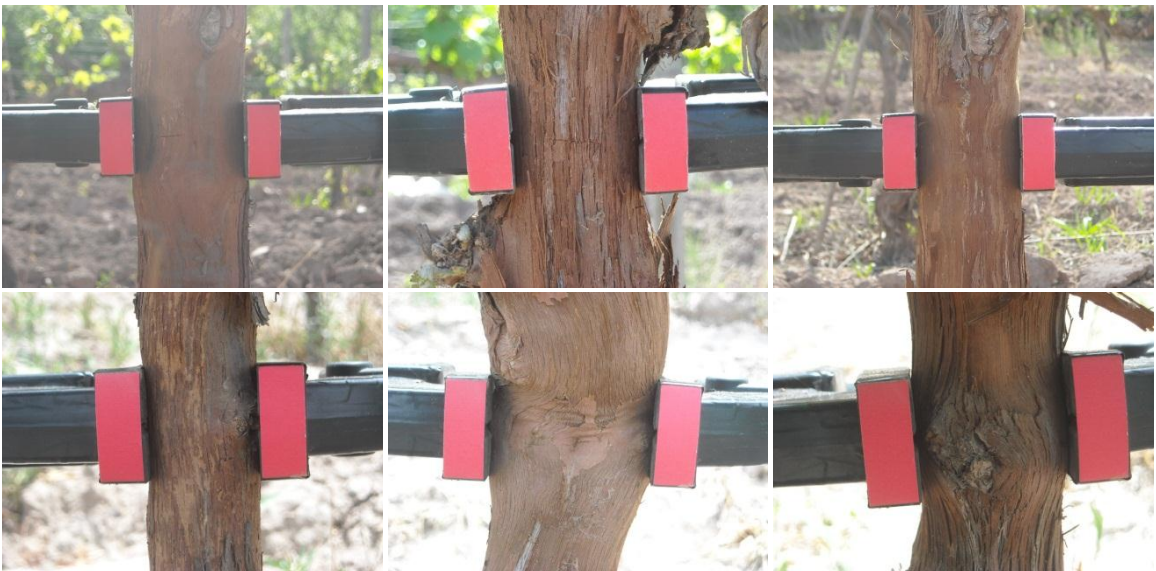

Figure 2. Examples of images used as training data to model the GMMs. The images were taken during sunny and cloudy days.

*Training Stage.* During the training stage, a set of manually segmented images are used to estimate two GMM distributions: a distribution for the spade pads and a distribution for the background. These distributions are used to model the probability that a pixel belongs to a certain class or object in the image (i.e. spade pads or background). This stage is conducted during the design of the algorithm, using images that are representative of the field conditions that may occur later during the measurement stage. In this work we used a well-mixed set of cloudy and sunny images, exemplified in figure 2. For training, these figures have been manually segmented into two classes: spade pads and background. The set of all pixels of all 8 images were separated into two datasets, one per class; with each datapoint containing the color information of the pixel. Experience from previous works (Pérez & Bromberg, 2012) shows that segmentation can be improved when the images are converted from RGB to LUV color-space (L=luminescence, U=saturation, V=hue angle), and the luminosity component discarded, i.e., using only the saturation and hue angle components. More details about color spaces can be found in (Jain et al., 1995; Szeliski, 2010). The two datasets (one per class) consist therefore on datapoints with two attributes, the U and V values of each pixel corresponding to the class of the dataset. Thus the resulting GMMs are 2D. Figure 3 illustrates this information as two 2D scatter plots over the UV space, one for the pixels in the background (right), and one for the pixels of the spade pads (left). In addition, two histograms are shown for each plot, one for the U component (on the top horizontal side) and one for the V component (on the right vertical side). These histograms show empirically the multimodal nature of the background, and the unimodal nature of the spade pads. This justifies using two modes GMM for modeling the spade pads color information, and three modes GMM for modeling the background color information.

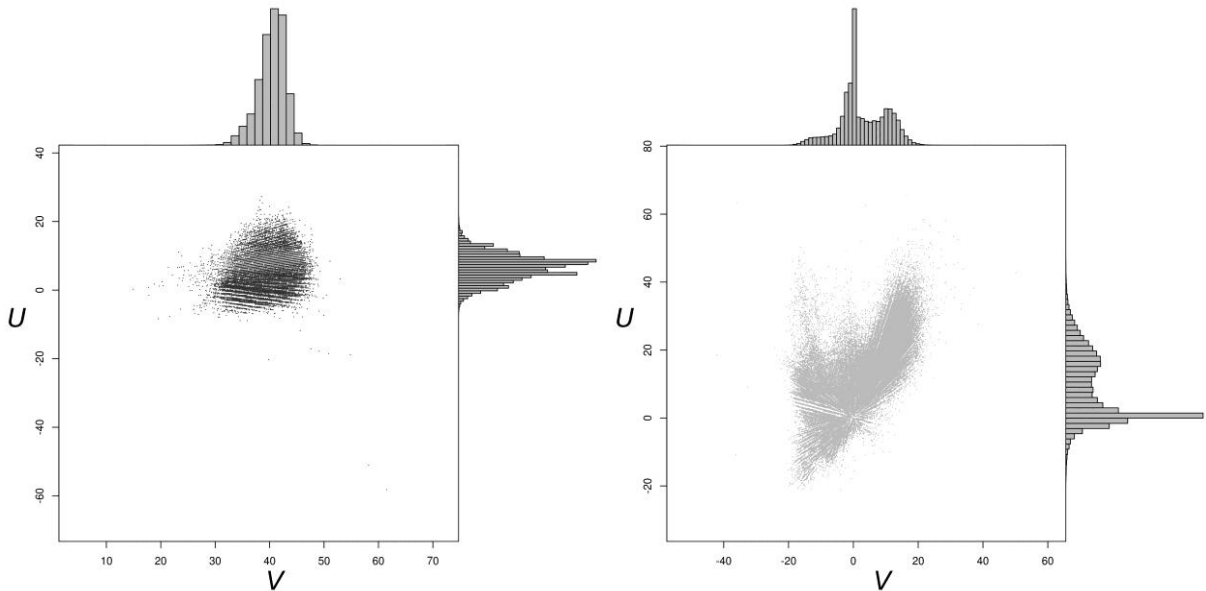

Figure 3. Scatter plot in the UV color-space for spade pads pixels (left) and the background pixels (right) of the image in figure 1b, together with the normalized histograms for U (top) and V (left) of each graph.

*Classification Stage.* Given the GMM estimated during the training stage, new images to be measures (i.e. not manually segmented) are autonomously segmented into interest classes: spade pads or background. This occurs in what is commonly called the classification stage, an algorithm with two steps: (i) convert new images from RGB to the LUV color space, discarding the L component; and (ii) discriminate each pixel as part of the spade pads class or the background class. The classification decision taken at the last step is carried out using the learned GMM, computing the probability of each pixel belonging to either class. The pixel is assigned to the class with highest probability.

## EXPERIMENTAL SETUP

In this section we describe the experimental setup used for confirming our main claim: that our *Image-based Measurement Method (IM)* for trunk diameter measurement produces results with precision equivalent to those obtained with the *Vernier Caliper Method (CM)*, in field conditions of climate and luminosity, while automating the measuring process and maintaining the same operational simplicity. The two methods are contrasted through their measurement errors. To obtain these errors we require the true value of the diameter. Although there may be several approaches for obtaining such value, we decided to use a non-standard, but nonetheless precise method: the *Manual Image-based Method (MIM)*, which matches exactly the IM method with the exception that the segmentation is performed manually using standard image editing software with extreme care. This method therefore avoids errors produced by the autonomous segmentation method, only introducing rounding errors caused by the resolution of the camera. Pixels in this work correspond to approximately 0.1 mm, equivalent to the 0.05 mm precision of a Vernier caliper. In the case of IM, this results in a very convenient error that measures the error of the segmentation. In the case of CM, this error highlights the potential manual errors I, II, or IV (c.f. table 1) caused by an operator, either due to some imperfection of their senses (potentially altered by the ambient context) or by their ability and experience in taking these type of manual measurements.

For a robust evaluation we compared the two methods over several images taken in field conditions of ambient luminosity (affected by climate) and different experiences of the operators, on real vineyards with vertical trellising system in the experimental fields of the Plant Physiology Laboratory, Department of Agricultural Sciences, National University of Cuyo (Lujan de Cuyo, Mendoza, Argentina). All the pictures were taken using a NIKON COOLPIX L16 compact commercial camera, in JPEG format with a resolution of 2048x1536 pixels (3 megapixels). For the caliper measurements we used a standard, plastic, Vernier caliper with precision of 0.5 mm. First, the measurement of the diameter was done by manipulating the caliper to set the magnitude and then this result was recorded on notepads or paper forms. Later, information was loaded in computer spreadsheets for its posterior processing and analysis.

**OPERATOR ERRORS EXPERIMENT**

In the first experiment the intent was to reproduce realistic scenarios for our operators. The luminosity conditions were variable, taking all pictures under sunny or cloudy conditions. Four operators were chosen and they were instructed to be extremely careful in their measurements (especially on the CM). The intent was to minimize the potential human error on these measurements, thus challenging our competitor IM. This scenario was designed to reproduce the most common conditions under which the CM is currently conducted in real-life.

For this experiment, each of the four operators conducted 7 rounds of CM over each of 30 grapevine plants, and 7 rounds of IM over the same plants. This totaled 840 measurements per method, 210 per operator, as well as a total of 28 rounds per method. The caliper and image measurements were taken in the same position of each plant's trunk in the location indicated by a previously positioned string as shown in figure 1b. Also, prior to the measurements, the bark was removed from each plant, around the target measurement position. By bark we refer to the rhytidome or dead tissue that forms around the trunk of a woody plant. In grapevine plants the bark is particularly thick, rough and breaks off in longitudinal strips, and can represent up to 10% of the trunk diameter. By removing the bark, the observed variability in data corresponds only to the errors of table 1. We report the mean and standard deviation of the absolute errors $|MM - MIM|$ and $|CM - MIM|$ of the IM and CM methods, respectively.

**LUMINOSITY CONDITIONS EXPERIMENT**

In the second experiment the intent was to challenge our image-based method by imposing extreme solar luminosity conditions. Solar luminosity is strongly affected by climate conditions such as cloudy skies, which may change radically, even during the duration of a measuring campaign (see figure 2). The experiment consisted in training the GMM using only images of one condition type (e.g., sunny), and testing its performance by segmenting images taken during another luminosity condition (e.g., cloudy). Then, the diameters obtained were contrasted against those obtained by matching conditions, in our example segmentation of sunny images over the sunny model. Specifically, the experiment consisted in two sets of 20 images each, the sunny set taken during a sunny day, and the cloudy set taken during a cloudy day. From each set, a subgroup of 4 images was separated randomly and used for training the corresponding GMMs $GMM_{Cloudy}$ and $GMM_{Sunny}$, leaving a total of 16 images on each set used later for testing the autonomous segmentation (denoted $Test_{Sunny}$ and $Test_{Cloudy}$ respectively). The two models and two 16 images sets were used in two crossed evaluations. One in which the 16 sunny images were segmented using the cloudy model, denoted $Test_{Sunny}|GMM_{Cloudy}$, and another where the 16 cloudy images were segmented using the sunny model, denoted $Test_{Cloudy}|GMM_{Sunny}$. To assess the precision of these segmentations, each case was contrasted against the case of matching conditions, i.e., the sunny set segmented against the sunny model (instead of the cloudy one), denoted $Test_{Sunny}|GMM_{Sunny}$, and the cloudy set segmented over the cloudy model (instead of the sunny one), denoted $Test_{Cloudy}|GMM_{Cloudy}$. In summary: $Test_{Sunny}|GMM_{Cloudy}$ tested against $Test_{Sunny}|GMM_{Sunny}$; and $Test_{Cloudy}|GMM_{Sunny}$ tested against $Test_{Cloudy}|GMM_{Cloudy}$.

## RESULTS

This section reports and discusses the results obtained for our two experiments. In both cases we report some aggregation of the errors of the IM and CM methods.

**OPERATOR ERRORS RESULTS**

The results for the operator errors experiment are reported here. We evaluated the mean and standard deviation of the absolute error over the 840 measurements of the all operators. The results show that the mean errors are smaller for IM, with 1.05 mm against the 1.34 mm of CM. The same holds for the standard deviations, with 0.95 mm for IM against 1.44 mm for CM. The smaller standard deviation implies that it is more probable for IM to have errors closer to the (smaller) mean. These results prove, at least empirically, that our method is equivalent (slightly better) to the manual caliper method. To conclude, we present a more detailed analysis of the outliers that produce the higher standard deviation of the CM method. For that, we computed a histogram that counts the number of errors for each method. For each possible error value (ranging from 0.0 to 1.8 mm), the histogram reports the number of measurements (out of the 840) that resulted in such an error (see figure 4). First we note that of all 840 cases, 62.2% of the CM results are smaller than the maximum mean 1.34 mm (i.e. CM mean), while IM has 7.2% more cases within that range, i.e., 69.4%. Also, all the cases of the IM method are smaller than 6.2 mm, while only 98.6% of the CM measurements fall within that interval, with the remaining 1.4% reaching errors of up to 18 mm. The 1.4% of 840 represents 12 measurements, which cannot be overlooked easily, and whose most probable cause is human error due to operator carelessness or fatigue.

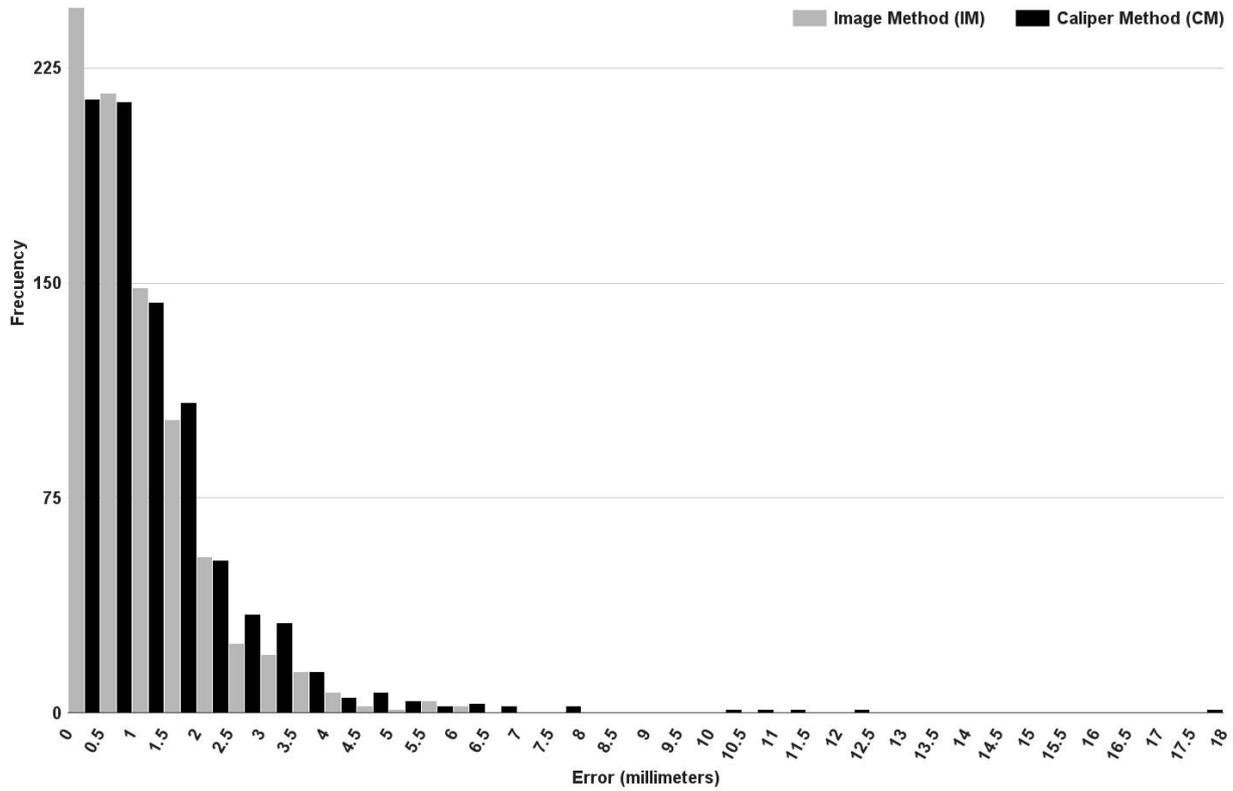

Figure 4. Histogram of errors for each method on each error range value to analysis of outliers.

We also analyzed the results with some more detail over each of the 7 individual measurement rounds conducted per operator for each method, were we report the error mean and standard deviation over the 30 measurements of each round. These results are shown in Figure 5, with one area curve per method, and were the results were sorted from highest to lowest mean error. The dark gray curve corresponds to IM while light gray corresponds to CM. The bars in each round point show the magnitude of the standard deviation. Given the imposed order, it is important to note that rounds labeled with the same number in IM and CM are not directly comparable with each other. However, the graph shows a clear improvement of IM over CM: the area under the IM curve is lower than the area under the CM curve, i.e. the possibility of errors is lower for rounds measured with IM than those measured with CM. For instance, the round with smaller error using CM (the number 28), is larger than the mean error of 17th round of IM, and therefore larger than the last 11 rounds of IM.

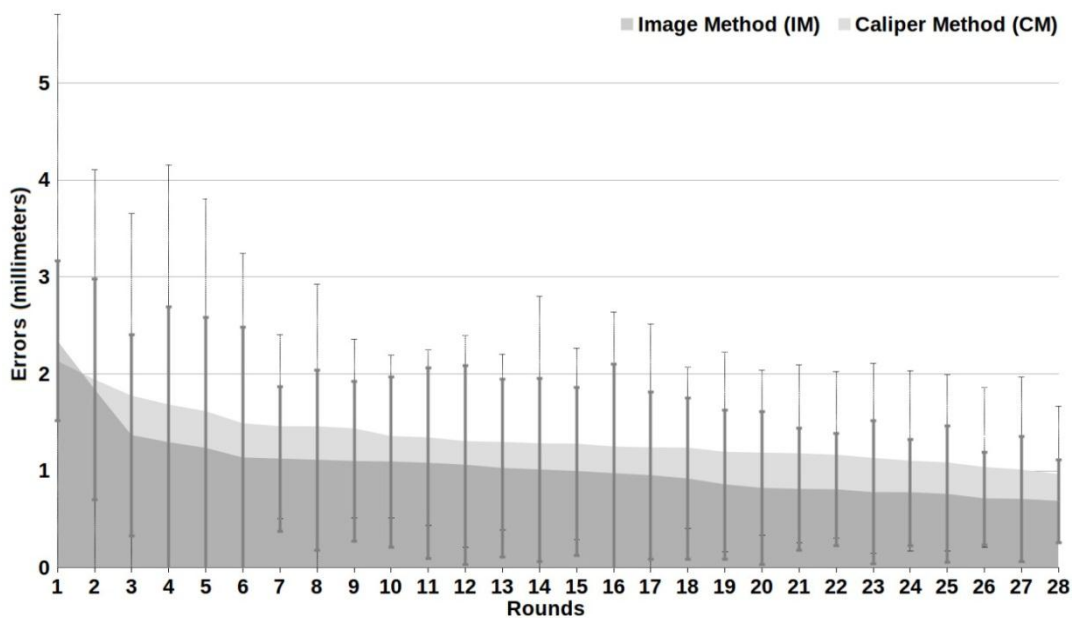

Figure 5. Mean error and its standard deviation for each of the 28 measurement rounds for the 30 plants, for both methods, sorted from highest to lowest mean error.

**LUMINOSITY CONDITIONS RESULTS**

The results for the luminosity conditions experiment are shown in table 2 (in millimeters). The table shows the comparison of the two cloudy cases $Test_{Cloudy}|GMM_{Sunny}$ vs. $Test_{Cloudy}|GMM_{Cloudy}$ (on top row) and the comparison of the two sunny cases $Test_{Sunny}|GMM_{Cloudy}$ tested against $Test_{Sunny}|GMM_{Sunny}$ (on the bottom row), for each of the 16 images, together with mean and standard deviation over the 16 images on the right. Both cases show a mean of approximately 0.45 mm, a small discrepancy considering the errors of IM (as shown in the previous experiment) are approximately 1 mm. Therefore, we can conclude that, at least empirically, the segmentation algorithm is robust to changes in luminosity caused by cloudy skies. This justifies using a single, mixed luminosity GMM for both luminosity conditions, as it was the case for the first experiment.

Table 2. Comparison of the cloudy and sunny cases by IM method. For each measurement obtained from 16 test images, it shows mean and standard deviation (in millimeters) of its differences (the measures for each case are omitted). Top row, results comparing the two cloudy cases $Test_{Cloudy}|GMM_{Sunny}$ vs. $Test_{Cloudy}|GMM_{Cloudy}$. Bottom row, results for the two sunny cases $Test_{Sunny}|GMM_{Cloudy}$ tested against $Test_{Sunny}|GMM_{Sunny}$.

| Method | MEAN *(SD)* |
|---|---|
| $Test_{Cloudy}|GMM_{Cloudy}$ vs. $Test_{Cloudy}|GMM_{Sunny}$ | 0.476 *(0.26)* |
| $Test_{Sunny}|GMM_{Sunny}$ vs. $Test_{Sunny}|GMM_{Cloudy}$ | 0.421 *(0.328)* |

# DISCUSSIONS

Different autonomous measurement methods have been introduced in the literature with the hope of mitigating the errors summarized in table 1. Mostly, their intent is not so much to increase the precision of individual measurements, as manual methods already present sufficient precision, but rather to mitigate the effect of manual or interpolation errors due to human fatigue or elevated costs of human labor. An autonomous system has the potential to reduce the required human intervention to a minimum during the measurement process. Our method, although not entirely autonomous as it requires fixing a quick clamp manually and taking a photograph, it requires little to none knowledgeable decision making, which makes it suitable to be operated by unknowledgeable, and thus lower stipend or more available personnel. Moreover, as we commented in the introduction, the quick clamp can be replaced by alternative devices with even lower labor costs. For example, it is possible to build a device cheaper and lighter than the quick clamp, so that it is gripped to the trunk at the beginning of the season to perform measurement campaigns throughout it. In this example scenario, the most arduous task, i.e. grip the clamp, is performed once in the season by knowledgeable personnel (e.g., researchers), and the measurement process is reduce to take a picture per plant and process it with our image segmentation method, that requires less to none expert knowledge.

Furthermore, it is possible to envision the possibility to use the proposed method to other types of fruit trees, such as apple, pear, peach, plum, and more; with only minor technical difficulties associated with building a suitable clamp to the characteristics of the new tree. While we have not assessed this proposal, it is expected that results will be as successful as those presented in this paper.

Also, this method presents the foundations for an embedded system within a mobile system that autonomously measures a large number of plants (e.g., unmanned land or air vehicles). The full automation could be achieved, for example, through the use of permanent cheap clamps attached to grapevine trunks. This autonomous mobile system should be able to identify the clamp in the scene, capture the image perpendicularly to the plane of spade pads, and finally perform the segmentation process and calibration. Such a system would allow massive measurements of the entire population of plants, with the consequent benefits in crop management operations.

It is worth noting some open problems around the proposed approach. Our method is still prone to error III of table 1, since plants are not identified autonomously, thus requiring a human to associate a photo ID with the plant. However, this is not difficult to automate. For example, one solution is to include QR codes (Liu, Gao, & Zhang, 2012) in the scene to identify automatically the plant of the image. Moreover, if the QR code is attached to the trunk by a permanent cheap clamp or mechanism placed previously, so as not to affect the growth of the trunk, the cost of the identification process could be amortized over several measurements (typical in agricultural settings) while the measurement process could be accelerated significantly. In addition, it may be embedded in an information system for the management of experiment campaigns that integrates the workflow for the whole agronomic experiment. Such a system could incorporate all the benefits of current technologies such as mobile photography and computing, GPS positioning, databases, web-based interfaces, and more (McBratney, Whelan, Ancev, & Bouma, 2005).

Finally, it is important to note that the errors incurred by the GMM algorithm during the automatic segmentation process, are given only over a *corrupted zone*, corresponding to edges of the spade pads in the images. In figure 1c the error by corrupted zone is approximately 9 pixels corresponding to 0.9 mm, but this value can vary depending of the images. Such error is caused by three main sources: imperfections of materials used in spade pads of the clamp;

presence of high brightness over the edges of spade pads; information loss due to JPEG compression algorithm. Thus, our approach may improve by simple technical enhancements such as better materials for the spade pads that better resists wear and tear, and using an uncompressed image format, further reducing segmentation errors.

## CONCLUSIONS

This paper has presented a high-precision and low-cost method based on Computer Vision for measurement of trunk diameter on grapevines from images acquired with a compact digital camera. This method is a contribution in the field of autonomous sensing of complex variables. In practice, generally, the trunk diameter is measured manually using Vernier calipers. This method involves human errors that can occur during measurement campaigns due to fatigue, haste or carelessness. There are several studies that address the problem of automating the measurement of trunk diameter trees. However, these solutions are not optimal: some are low cost but low precision (or equivalently high failing rate), and others are high precision but high cost. The experimental analysis for our method shows that it is slightly more accurate than the manual method based on caliber. An important advantage over other authors named in this work, is that the technology required to implement this method is cheap and probably any agronomist already has some of the main elements: a standard desktop computer and a low-resolution digital camera (3 megapixels). In addition, our method has two advantages over the manual method: it avoids mistakes I, II, and IV, summarized in table 1; and is susceptible to be fully automated, with the potential to reduce errors III and V.

**ACKNOWLEDGEMENTS**

This work was funded by the scholarship program of the National Technological University (UTN), the National Council of Scientific and Technical Research (CONICET) of Argentina, and the National Fund for Scientific and Technological Promotion (FONCyT) of Argentina. We thank the Laboratory of Plant Physiology, Department of Agricultural Sciences, UNCuyo, for offering their vineyards to capture the images used in this work.

## REFERENCES

Bishop, C. M. (2009). *Pattern recognition and machine learning*. Springer.

Brunelli, R. (2009). *Template Matching Techniques in Computer Vision: Theory and Practice*. Wiley.

Castelan-Estrada, M., Vivin, P., & Gaudillière, J. P. (2002). Allometric Relationships to Estimate Seasonal Aboveground Vegetative and Reproductive Biomass of Vitis vinifera L. *Annals of Botany*, *89*(4), 401–408.

Causton, D. R. (1985). Biometrical, structural and physiological relationships among tree parts. *Attributes of Trees as Crop Plants. Edited by MGR Cannel and JE Jackson. Titus Wilson & Son Ltd.. Cumbria, Great Britain*, 137–159.

Cressie, N. (1990). The origins of kriging. *Mathematical Geology*, *22*(3), 239–252.

Hastie, T., Tibshirani, R., & Friedman, J. H. (2009). *The Elements of Statistical Learning: Data Mining, Inference, and Prediction*. Springer.

Jain, R., Kasturi, R., & Schunck, B. G. (1995). *Machine vision* (Vol. 5). McGraw-Hill New York.

Jutila, J., Kannas, K., & Visala, A. (2007). Tree measurement in forest by 2D laser scanning. In *International Symposium on Computational Intelligence in Robotics and Automation, CIRA 2007* (pp. 491–496). Jacksonville, Florida, U.S.A: IEEE.

Kan, J., Li, W., & Sun, R. (2008). Automatic measurement of trunk and branch diameter of standing trees based on computer vision. In *3rd IEEE Conference on Industrial Electronics and Applications, ICIEA 2008* (pp. 995–998). Singapore, Singapore: IEEE.

Liu, Y.-C., Gao, H., & Zhang, X. (2012). Development and Application of Mobile Traceability Data Construction for Agriculture. In *8th Asian Conference for Information Technology in Agriculture and World Conference on Computer in Agriculture*. Taipei City, Taiwan.

MacAdam, J. W. (2013). *Structure and Function of Plants*. John Wiley & Sons.

McBratney, A., Whelan, B., Ancev, T., & Bouma, J. (2005). Future directions of precision agriculture. *Precision Agriculture*, *6*(1), 7–23.

Niklas, K. J. (1994). *Plant Allometry: The Scaling of Form and Process*. University of Chicago Press.

Niklas, K. J. (1995). Size-dependent allometry of tree height, diameter and trunk-taper. *Annals of Botany*, *75*(3), 217–227.

Omasa, K., Hosoi, F., Uenishi, T. M., Shimizu, Y., & Akiyama, Y. (2008). Three-dimensional modeling of an urban park and trees by combined airborne and portable on-ground scanning LIDAR remote sensing. *Environmental Modeling & Assessment*, *13*(4), 473–481.

Pérez, D. S., & Bromberg, F. (2012). Segmentación de imágenes en viñedos para la medición autónoma de variables vitícolas. In *XVIII Congreso Argentino de Ciencias de la Computación*.

Szeliski, R. (2010). *Computer Vision: Algorithms and Applications*. Springer.

Tetuko, J., Tateishi, R., & Wikantika, K. (2001). A method to estimate tree trunk diameter and its application to discriminate Java-Indonesia tropical forests. *International Journal of Remote Sensing*, *22*(1), 177–183.

Thamrin, N. M., Arshad, N. H. M., Adnan, R., Sam, R., Razak, N. A., Misnan, M. F., & Mahmud, S. F. (2013). Tree diameter measurement using single infrared sensor for non-stationary vehicle context in agriculture field. In *4th Control and System Graduate Research Colloquium (ICSGRC)* (pp. 38–42). Shah Alam, Malaysia: IEEE.